# Fine-Tuning MedGemma for Clinical Captioning to Enhance Multimodal RAG over Malaysia CPGs

Lee Qi Zun
*Qmed AI Team*
*Qmed Asia*
Kuala Lumpur, Malaysia
Email: jacky@qmed.asia

Mohamad Zulhilmi Bin Abdul Halim
*Qmed AI Team*
*Qmed Asia*
Kuala Lumpur, Malaysia
Email: m.zulhilmi.a.halim@gmail.com

Goh Man Fye
*Qmed AI Team*
*Qmed Asia*
Kuala Lumpur, Malaysia
Email: manfye@qmed.asia

***Abstract*—Retrieval-Augmented Generation (RAG) systems are essential for providing fact-based guidance from Malaysian Clinical Practice Guidelines (CPGs), but their effectiveness with image-based queries is limited, as general Vision-Language Model (VLM) captions often lack clinical specificity and factual grounding. This study proposes and validates a framework to specialize the MedGemma model to generate high-fidelity captions that can serve as superior queries. To overcome data scarcity, we employ a knowledge distillation pipeline to create a synthetic dataset across dermatology, fundus, and chest radiography domains, then fine-tune MedGemma using the parameter-efficient QLoRA method. Performance was rigorously assessed via a dual framework measuring both classification accuracy and, through a novel application of the RAGAS framework, caption faithfulness, relevancy and correctness. The fine-tuned model demonstrated substantial improvements in classification while RAGAS evaluation also confirmed significant gains in caption faithfulness and correctness, validating the model's ability to generate more reliable, factually grounded descriptions. This work establishes a robust pipeline for specializing medical VLMs and validates the resulting model as a high-quality query generator, laying the essential groundwork for enhancing multimodal RAG systems for evidence-based clinical decision support.**

## I. INTRODUCTION

In Malaysia, evidence-based clinical practice is strongly advocated by the Ministry of Health through the development and dissemination of national Clinical Practice Guidelines (CPGs) [1] [2]. These documents form the cornerstone of standardized patient care, providing systematically developed recommendations. As the healthcare sector undergoes a digital transformation, there is a growing imperative to make the rich knowledge within these guidelines more accessible. This has led to the exploration of large language models (LLMs) as a tool for clinical decision support. However, the deployment has revealed a critical reliability issue which is the tendency of LLMs to generate non-factual medical information [3] [4]. To mitigate this risk, Retrieval-Augmented Generation (RAG) has emerged as an essential architecture. By grounding model outputs in a verifiable knowledge base such as Malaysian CPGs, a RAG system can deliver factual and trustworthy clinical guidance [5] [6] [7] [35].

While CPGs provide an essential textual foundation, the clinical workflow is inherently multimodal, with medical image interpretation being a critical component of diagnosis and treatment planning. The advent of Vision-Language Models (VLMs) [8], which have significantly advanced the ability to associate images with descriptive text, presents a promising pathway for integrating such visual data into an RAG system. Despite these advancements, a significant performance gap remains. Generic VLMs, primarily trained on natural image datasets, struggle to interpret the medical imaging, often failing to capture radiologic image intricacies [9] [10]. This often results in descriptions that are inaccurate or clinically irrelevant, a deficiency that critically limits their utility in high-stakes medical contexts.

To bridge this performance gap, medically-aware foundation models like MedGemma provide a robust starting point [11]. As a vision-language model built on the Gemma 3 architecture [12], its visual acuity is powered by MedSigLIP, a vision encoder specifically tuned for medical data from the high-performing SigLIP model. This specialized encoder gives MedGemma visual interpretation capabilities that rival or even exceed many task-specific models, enabling advanced medical reasoning across both images and text [11]. While this powerful foundation is a significant advantage, its effectiveness within a multimodal RAG system is ultimately contingent on its output [13]. For the system to function optimally, the query generated from an image must be of the highest fidelity. A generic caption is therefore insufficient; a precise, clinically-aware description is needed to retrieve the most relevant guidance from the CPG knowledge base.

To address these limitations, this research focuses on enhancing MedGemma's captioning capabilities through domain-specific fine-tuning. In this case, medical images captioning refers to the ability to understand the content of medical images and delivering accurate descriptions [14] [15].The core idea is that a highly specialized medical caption can serve as a superior, meaningful query for the multimodal RAG system, unlocking more accurate information from the CPGs. A central challenge to this approach is the scarcity of large-scale, high-quality image-caption pairs for specialized medical domains. This study overcomes this obstacle by employing knowledge distillation from a state-of-the-art teacher model (Open AI GPT-5). Recent research has shown that GPT-5 delivered substantial gains in multimodal medical reasoning [16] [17]. By leveraging such a model to generate and then filter synthetic captions from existing classification datasets, a high-quality fine-tuning corpus was created.

Therefore, this paper proposes and evaluates a framework to enhance clinical decision support by specializing MedGemma for high-fidelity query generation in a multimodal RAG system. The study first details the development of a novel training dataset for dermatology, fundus, and chest radiography using a knowledge distillation pipeline. It then describes the fine-tuning of the MedGemma-4B-IT model to produce clinically-specific and uncertainty-aware captions. Finally, the fine-tuned model's performance is rigorously evaluated on both its classification accuracy and the clinical fidelity of its generated captions, demonstrating its downstream utility.

## II. Related Works

The application of Vision-Language Models (VLMs) in medicine has been primarily driven by the need to overcome the domain gap of general-purpose models, which often fail to interpret specialized clinical imagery [9] [10]. This led to the development of medically-aware foundation models such as Med-PaLM, Med-PALM 2, LLaVA-Med, and notably, MedGemma [11] [26] [27] [28]. These models are pre-trained on large-scale medical corpora, such as PMC-15M, MIMIC-CXR, and CheXpert, equipping them with a strong prior for clinical understanding. A review of the literature shows that the predominant application and evaluation of these powerful models have centered on discriminative tasks. For example, their performance has been extensively benchmarked on medical image classification and Visual Question Answering (VQA), where the goal is to produce a discrete label or a short-form answer [29]. While these studies have firmly established the state-of-the-art capabilities of VLMs for classification, the shift from producing a single correct label to generating reliable, free-form text introduces a host of new challenges in factuality and evaluation. [3] [4].

Medical image captioning traditionally aims to generate comprehensive narratives that mimic full radiology reports [30] [31]. This paradigm, however, often prioritizes linguistic similarity over clinical precision. The central research challenge, therefore, is not the ability to generate fluent text, but the lack of focus on producing verifiable and visually grounded captions. A significant challenge for generative VLMs is the issue of hallucinations, where models generate outputs that are not grounded in the provided images or are inconsistent with established knowledge [32] [33]. In a medical context, such failures can lead to dangerously inaccurate diagnostic information.

Evaluating the outputs of medical VLMs has traditionally followed two main pathways, corresponding to their discriminative and generative capabilities. For discriminative tasks, performance is commonly measured using standard classification metrics such as accuracy, precision, recall, and the F1-score [29]. In the context of report generation, these metrics are often referred to as "clinical efficacy metrics" and are computed by applying a labeler like CheXpert to both the generated and reference reports to compare the presence of clinical findings. For generative tasks like report generation and captioning, the quality of the output has typically been assessed using text-similarity metrics like Bilingual Evaluation Understudy (BLEU) and Recall-Oriented Understudy for Gisting Evaluation (ROUGE) [29] [30]. These metrics evaluate the overlap of n-grams between the model-generated text and a human-generated reference. Lexical overlap scores from BLEU and ROUGE often fail to capture the clinical accuracy or semantic correctness of a generated report, and a model can achieve a high score while producing a factually incorrect or clinically misleading statement.

The preceding review of the literature reveals a critical and consolidated research gap. While medical VLMs have proven highly effective for classification tasks, a significant challenge remains in ensuring the factual and visual grounding of their generative outputs. The problem of "hallucinations" is a major barrier to their safe deployment in clinical settings [32] [33]. This challenge is compounded by the historical reliance on evaluation methodologies that are ill-suited for high-stakes medical applications. The inadequacy of lexical metrics like BLEU and ROUGE to measure clinical accuracy means that a core aspect of model performance is often overlooked. Consequently, there is a clear need for research that not only reframes the objective of medical captioning from linguistic mimicry to factual reporting but also employs a more sophisticated evaluation framework capable of quantitatively assessing both clinical accuracy and faithfulness to the source image.

## III. Methodology

This section details the systematic framework developed to fine-tune and evaluate the MedGemma model for high-fidelity medical image captioning. The methodology is designed to be a replicable pipeline, starting from raw data acquisition and culminating in a rigorous, multi-faceted evaluation of the final model's output. Each stage of the process is engineered to address the specific challenges of generating clinically accurate and visually faithful medical descriptions.

The core of the methodology is a four-stage pipeline designed to create a specialized captioning model and rigorously validate its performance as shown in Fig. 1. The process flows from data collection to final evaluation, with each step building upon the last to ensure the final model is both robust and reliable. The following subsections describe each stage in detail.

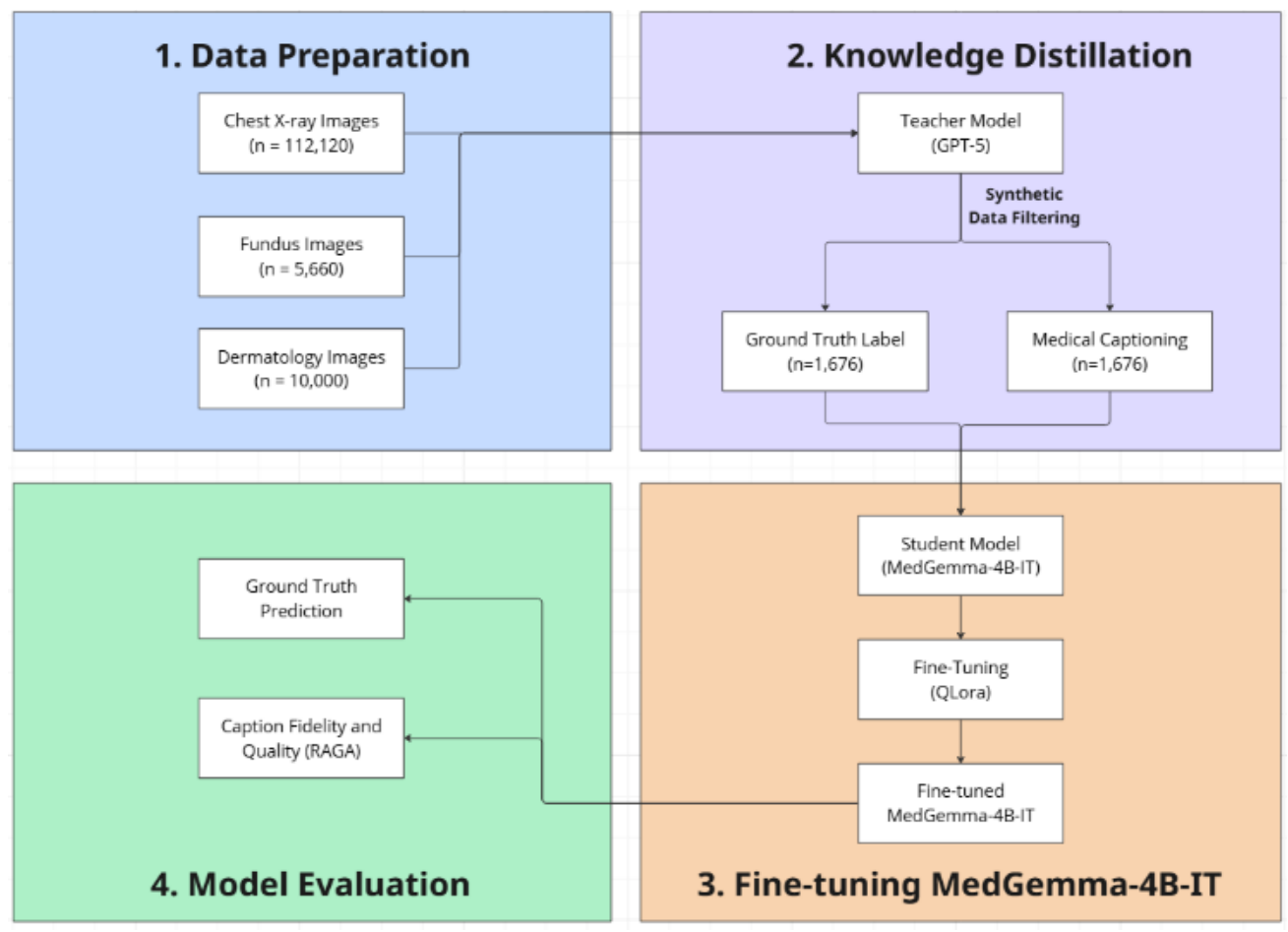

Fig. 1. Proposed four-stage pipeline in finetuning MedGemma-4B-IT.

### A. *Dataset Preprocessing*

The source images for this study were drawn from three distinct, publicly available, and clinically-annotated medical imaging collections to ensure multimodal diversity and data quality. Fundus photography images were sourced from the Asia Pacific Tele-Ophthalmology Society (APTOS) 2019 Blindness Detection dataset [22], which contains high-resolution retinal images graded for diabetic retinopathy across five severity levels (Grade 0 to Grade 4). For the Chest X-ray (CXR) modality, images were utilized from the National Institutes of Health (NIH) Chest X-Ray Dataset [23], comprising radiographs annotated with 14 common thoracic pathologies and one normal class. Finally, dermatological images were obtained from the Human Against Machine with 10,000 training images (HAM10000) dataset [24], a collection of multi-source dermoscopic images classified into seven distinct categories of pigmented skin lesions. Collectively, these large-scale datasets provided the diverse pool of images used as input for the iterative knowledge distillation and filtering pipeline, from which the final balanced training corpus was constructed.

### B. Knowledge Distillation

In this phase, this study employed a knowledge distillation pipeline to generate a rich, synthetic dataset for fine-tuning. This approach leverages the advanced capabilities of a large, state-of-the-art "teacher" model to create high-quality training data for a smaller, more specialized "student" model, MedGemma. The teacher model, GPT-5, is chosen for this task for its demonstrated state-of-the-art performance in complex, multimodal clinical reasoning tasks [16] [17]. Its ability to generate nuanced, clinically accurate interpretations from medical images makes it a suitable proxy for human expert annotation, providing the "ground truth" for our synthetic dataset.

#### a) Teacher Generation

The knowledge distillation process began with an iterative generation and filtering procedure designed to create a high-quality and class-balanced synthetic dataset. Instead of using a fixed, pre-sampled set of images, images from the three large-scale source datasets (APTOS, NIH Chest-Xray, and HAM10000) were systematically fed into the GPT-5 endpoint. To ensure the generated data was consistent and comprehensive, the model was instructed to produce a structured output in JSON format for each image.

This JSON object contained two critical key-value pairs with a specific nested structure for the description, as detailed in Table I. The prompt also included guardrail instructions, directing the model to act as an interpretive tool only and to refrain from providing management advice.

TABLE I. STRUCTURE OF THE GENERATED JSON OUTPUT

| Key | Description |
| --- | --- |
| Prediction | A single, canonical class label corresponding to the ground-truth diagnosis. |
| Description | A structured analysis containing the sections.<br>**1. IMAGE TYPE**: The imaging modality.<br>**2. ANATOMICAL REGION**: The specific body part depicted.<br>**3. KEY FINDINGS**: Objective observations of pathologies.<br>**4. CLINICAL SIGNIFICANCE**: Relevant diagnostic implications. |

This structured format was designed to yield rich, multi-faceted data points that go beyond simple labels, providing both a classification target and a detailed descriptive narrative for each image. The same set of images was also processed by the base MedGemma-4B-IT model to generate its pre-fine-tuning predictions and descriptions.

#### b) Synthetic Data Filtering

The outputs generated during the iterative Teacher Generation phase underwent a rigorous filtering process to ensure the classification performance and overall quality of the training corpus. For each image successfully processed, the outputs from the teacher model were associated with their corresponding ground-truth class labels.

The primary filtering criterion was classification correctness. An image-caption pair was only retained if the prediction key in its generated JSON object exactly matched the ground-truth class label. This critical step ensures that the fine-tuning process is not contaminated with factually incorrect examples, preventing the propagation of errorsto the student model, MedGemma.

#### c) Filtering Outcomes

The iterative generation combined with the stringent, ground-truth-based filtering process yielded a refined, high-fidelity, and class-balanced dataset suitable for fine-tuning. The process concluded once a sufficient number of correct samples were collected for each targeted class across the three domains. For the chest-xray dataset, this resulted in 500 samples distributed evenly across 10 distinct classes which five of the original 15 classes were excluded due to insufficient yields during the filtering phase. The Dermatology dataset comprised 676 samples, balanced across its seven lesion categories. Similarly, the Fundus dataset contained 500 samples, balanced across the five grades of diabetic retinopathy. This meticulous filtering and balancing approach yielded a final, high-quality fine-tuning corpus containing a total of 1,676 image-caption pairs. This dataset formed the exclusive basis for the subsequent partitioning and fine-tuning stages.

#### d) Dataset Partitioning for Training and Evaluation

Following the filtering process, the finalized corpus of 1,676 high-quality image-caption pairs was partitioned into three independent subsets for model training, validation, and final evaluation. A conventional 70:20:10 split was employed, resulting in a training set of 1,173 samples, a validation set of 335 samples, and a test set of 168 samples. The training set was used exclusively to update the model's parameters during the fine-tuning process, while the validation set was utilized to monitor for overfitting and guide hyperparameter selection.

The test set, held out from all training and tuning procedures, was reserved for the final, unbiased assessment of the model's performance. The composition of this 168-sample test set reflects the distribution of the filtered data, comprising 50 chest x-ray images, 68 dermatology images, and 50 fundus images. This partitioning ensures a rigorous and fair evaluation of the model's generalization capabilities across the three distinct medical domains.

### C. Finetuning MedGemma

This section outlines the technical details of the fine-tuning procedure used to specialize the MedGemma-4B-IT model. The process was designed to be memory-efficient while effectively teaching the model to generate structured, high-fidelity clinical captions from medical images.

#### a) Parameter-Efficient Fine-Tuning (PEFT) Strategy

The base model for this study was MedGemma-4B-IT, sourced from the Hugging Face Hub. Given the significant computational resources required to fine-tune a 4-billion-parameter model, a Parameter-Efficient Fine-Tuning (PEFT) strategy was employed. Specifically, Quantisation-aware

Low-Rank Adaptation (QLoRA) was selected to drastically reduce memory overhead while maintaining high model performance [25] [34].

*b) Instruction Tuning and Data Formatting*

The fine-tuning process utilized an instruction-based format to teach the model its specific task. Each sample in the dataset was structured into a conversational format consisting of a system prompt, a user prompt, and an assistant response.

A persistent system prompt established the model's persona as a "specialist clinician and image interpreter," and strictly mandated that all outputs conform to the JSON structure detailed in Table I. For each training instance, the model was provided with two inputs, the medical image and a general query prompting an interpretation of its findings. The supervised learning signal was the corresponding ground-truth JSON object from the distilled dataset, which functioned as the target assistant response for the model to learn and replicate. This instruction-tuning setup guides the model to learn not only the content of the clinical captions but also the precise, structured format required for the output.

*c) Training Configuration and Implementation*

The model was fine-tuned for 10 epochs using the Hugging Face SFTTrainer from TRL library. Training was configured with a per-device batch size of 4 and gradient accumulation steps of 4, resulting in an effective batch size of 16. The AdamW optimizer was used with a learning rate of $2\times10^{-4}$ and a linear learning rate scheduler with a warmup ratio of 0.03. To further conserve memory, gradient checkpointing was enabled. The loss function was the standard Cross-Entropy Loss, calculated on the model's next-token predictions. To ensure the model only learned from the target text, tokens corresponding to the input prompt, padding, and the image itself were masked out and ignored during the loss computation.

### *D. Dual-Phase Evaluation*

To comprehensively assess the performance of the fine-tuned MedGemma model, a multi-faceted evaluation framework was designed. This framework provides a holistic view of model improvements by assessing two key areas which include the concordance of its predictions with ground-truth labels, and the factual quality and visual groundedness of its generated captions. The evaluation was conducted on the held-out test set, comparing the performance of the fine-tuned model against the baseline, pre-trained MedGemma-4B-IT model.

*a) Classification Assessment*

This component of the evaluation quantitatively assesses the model's ability to correctly classify medical images. For each image in the held-out test set, the prediction key was extracted from the model's generated JSON output and compared against the true classification label. To gauge overall performance, standard Accuracy and Balanced Accuracy were used to measure the total proportion of correct predictions and mitigate the effect of class imbalance by averaging the recall of each class.

Next, the macro-averaged Precision, Recall, and F1-Score were also calculated to compute the metrics independently for each class before taking their unweighted mean. This ensures that each class contributes equally to the final score, providing an unbiased assessment of the model's ability to classify both common and rare categories.

*b) Caption Quality Assessment*

Beyond discrete classification, it is critical to evaluate the quality and factual groundedness of the generated free-form description. Traditional NLP metrics like BLEU or ROUGE are ill-suited for this, as they measure lexical overlap rather than clinical accuracy. Therefore, this study employs the RAGAS framework to quantitatively measure the quality of the generated text.

For this evaluation, the components of the RAGAS framework were mapped as follows. The high-quality teacher-generated description from the knowledge distillation phase served as both the "context" and the "ground_truth". The description generated by the baseline or fine-tuned MedGemma was treated as the "answer". This setup allows for a rigorous assessment of how well the fine-tuned model has learned to replicate the factual content and style of the teacher model.

The quality of each generated caption was scored across three key dimensions. Faithfulness was measured to ensure the model's description was factually consistent with and contained only information present in the teacher-provided context. Answer Relevancy was used to assess how well the caption addressed the core task of describing key findings. Finally, Answer Correctness was calculated to evaluate the factual alignment of the generated caption against the "ground_truth" description from the teacher model. This multi-metric approach provides a robust method for validating the success of the knowledge distillation process.

## IV. Technical Experiment

The experiments were conducted on the Microsoft Azure cloud platform, utilizing a high-performance instance equipped with NVIDIA A100 GPUs (80 GiB memory) and a 960 GB NVMe SSD for efficient data handling. The implementation was built on a Python 3.8 environment with the PyTorch (v2.7.1) deep learning framework, configured with CUDA 11.8 for GPU acceleration. The fine-tuning process was orchestrated using the Hugging Face ecosystem, leveraging transformers (v4.56.2) for model and processor handling, peft (v0.17.0) and bitsandbytes (v0.48.0) for the QLoRA implementation, and trl (v0.20.0) for the supervised fine-tuning trainer. For the evaluation phase, classification metrics were calculated using scikit-learn (v1.2.2) while caption fidelity was assessed using the ragas library (v0.3.6).

## V. Results

This section presents the empirical results of the study, quantitatively evaluating the performance of the fine-tuned MedGemma model against the baseline, pre-trained MedGemma-4B-IT model. The findings are organized into two key areas, corresponding to the evaluation framework detailed in the Methodology section. First, the classification performance of both models is assessed using a comprehensive suite of classification metrics. Second, the fidelity and quality of the generated captions are analyzed using the repurposed RAGAS framework. All results reported were generated using the held-out test set to ensure an unbiased assessment of model performance.

*a) Classification Assessment*

The classification performance of the fine-tuned MedGemma model was rigorously compared against the baseline MedGemma-4B-IT model on the held-out test set.

The comprehensive results, broken down by dataset, are presented in Table II.

TABLE II. GROUND-TRUTH CONCORDANCE ASSESSMENT ON TEST SET.

| Dataset | Model | Accuracy | Precision | Recall | F1 |
|---|---|---|---|---|---|
| Fundus (n=50) | Base | 0.5200 | 0.5368 | 0.6755 | 0.5368 |
| | Fine-Tuned | **0.6200** | **0.6674** | **0.6675** | **0.6674** |
| Dermatology (n=68) | Base | 0.0882 | 0.0813 | 0.0687 | 0.0711 |
| | Fine-Tuned | **0.4265** | **0.4870** | **0.4546** | **0.4870** |
| Chest-Xray (n=50) | Base | 0.4200 | 0.4908 | **0.6650** | 0.4908 |
| | Fine-Tuned | **0.5200** | **0.5558** | 0.5757 | **0.5558** |

The results of the classification performance assessment indicate a consistent improvement for the fine-tuned MedGemma model over its baseline counterpart, particularly in its ability to handle complex, multi-class domains.

The most significant performance gain was observed in the dermatology dataset. The fine-tuned model achieved an accuracy of 0.4265 from a very low baseline accuracy of just 0.0882. This substantial relative improvement is mirrored in the macro F1-score, which surged from 0.0676 to 0.4166, demonstrating that the fine-tuning process successfully imbued the model with the specialized knowledge required to distinguish between the seven different lesion types, a task the baseline model largely failed at.

For the fundus photography and chest x-ray datasets, the fine-tuned model also showed consistent gains in key metrics like Accuracy and Balanced Accuracy. While the improvement in the chest x-ray domain was more moderate, likely attributable to the smaller volume of training data available for this modality, the overall trend confirms the efficacy of the fine-tuning process. The comprehensive improvements provide strong quantitative evidence that domain-specific adaptation was highly effective in enhancing the model's classification capabilities.

*b) Caption Quality Assessment*

The fidelity and quality of the generated captions were quantitatively assessed using the RAGAS framework to evaluate how well the fine-tuned model learned from the high-quality, knowledge-distilled data. The evaluation was performed on the test set, comparing the fine-tuned MedGemma model against the baseline. Table III presents the mean scores for faithfulness, answer relevancy, and answer correctness, aggregated by medical domain.

TABLE III. CAPTION FIDELITY AND QUALITY ASSESSMENT USING RAGAS.

| Dataset | Model | Faithfulness | Answer Relevancy | Answer Correctness |
|---|---|---|---|---|
| Fundus (n=50) | Base | 0.2996 | 0.4426 | 0.4136 |
| | Fine-Tuned | **0.5662** | **0.4533** | **0.6213** |
| Dermatology (n=68) | Base | 0.3166 | 0.3800 | 0.2836 |
| | Fine-Tuned | **0.4467** | **0.4833** | **0.5605** |
| Chest-Xray (n=50) | Base | 0.3970 | 0.4890 | 0.4643 |
| | Fine-Tuned | **0.5331** | **0.5563** | **0.5774** |

The results from the RAGAS evaluation clearly indicate that the fine-tuned MedGemma model generates substantially higher-quality clinical descriptions than the baseline model. Across all three domains, the fine-tuned model achieved superior scores in all evaluated metrics, with the most significant improvements consistently seen in faithfulness and answer correctness.

The fine-tuned model's ability to produce factually grounded descriptions is best demonstrated by the faithfulness scores, which saw dramatic improvements. The fundus dataset showed the largest relative gain, with its score increasing by nearly 90% from a baseline of 0.2996 to 0.5662. This indicates that the fine-tuning process was highly effective at reducing model "hallucinations" and teaching the model to generate descriptions that are more verifiably supported by the context provided by the teacher model.

Similarly, the substantial gains in answer correctness, most notably the near-doubling of the score in the dermatology dataset which is from 0.2836 to 0.5605 has further confirmed that the fine-tuned model successfully learned to align its outputs with the factual content and clinical terminology of the high-quality synthetic data. While the improvements in answer relevancy were more modest across the board, the primary uplift in faithfulness and correctness provides strong quantitative evidence that the fine-tuned model is not only a better classifier but also a more reliable and factually accurate narrator of medical findings. This validates the success of the knowledge distillation and fine-tuning pipeline.

## VI. DISCUSSION

This study successfully demonstrated that a targeted, domain-specific fine-tuning process can substantially elevate the performance of a generalist medical foundation model like MedGemma. The empirical results confirm that this approach not only enhances classification capabilities but also represents a critical step toward generating high-fidelity medical captions suitable for downstream clinical applications. The primary finding is that our fine-tuning methodology, which pairs knowledge distillation with parameter-efficient techniques, effectively imbues the model with specialized clinical expertise. This was most evident in the dermatology dataset, where accuracy surged from a near-random baseline of 0.0882 to 0.4265, with the F1-score showing a dramatic rise. The consistent improvements in metrics like Balanced Accuracy across the fundus and chest-xray datasets further underscore the model's enhanced, more robust understanding of the different disease categories.

Beyond these classification improvements, the evaluation of caption fidelity using the RAGAS framework validates the model's enhanced generative quality. The fine-tuned model demonstrated marked and consistent improvements in faithfulness and answer correctness across all domains. The most significant relative gains were seen in the fundus dataset for faithfulness which is nearly doubling from 0.2996 to 0.5662 and the dermatology dataset for answer correctness is also nearly doubling from 0.2836 to 0.5605. This quantitatively demonstrates that the fine-tuning process significantly reduced the model's tendency for "hallucination," making it a more reliable narrator of visual evidence. The parallel rise in answer correctness scores confirms the success of the knowledge distillation pipeline, as the model learned to accurately replicate the factual content of the high-quality teacher-generated data.

Ultimately, the goal of enhancing the model's classification accuracy and caption fidelity is to create a superior input for a downstream clinical decision support system. The enhanced medical captioning capability of the fine-tuned MedGemma is specifically intended to serve as a high-fidelity, structured query for a multimodal RAG system operating on Malaysian Clinical Practice Guidelines (CPGs). A precise and factually grounded caption is hypothesized to be far more effective for retrieving relevant, context-specific guidance. This work therefore validates the foundational component of that larger vision: the ability to reliably transform a raw medical image into a machine-readable, clinically-aware textual summary.

## VII. Limitation and Future Works

Despite the promising results, this study has several limitations that provide clear directions for future research. Firstly, the performance of the fine-tuned model is inherently capped by the quality and accuracy of the teacher model used for knowledge distillation. Any errors or biases present in the teacher model's outputs can be propagated to the student model. This was exemplified by the chest x-ray dataset, where the teacher model's lower initial accuracy directly resulted in a smaller and potentially less diverse training corpus, constraining the final performance.

Secondly, the study's scope was limited to three medical domains. While the framework proved effective for dermatology, fundus photography, and chest radiography, further research is required to validate its generalizability to other imaging modalities with different characteristics, such as CT scans, MRIs, or histopathology slides.

The primary avenue for future work is the integration of this specialized captioning model into the end-to-end multimodal RAG system it was designed to support. The clear next step is to use the high-fidelity captions as queries to retrieve information from the Malaysian CPGs and empirically measure the performance of the full pipeline. This will involve evaluating the RAG system's ability to provide accurate, relevant, and contextually appropriate clinical guidance based on an initial image query. Success in this next phase would complete the proposed framework, delivering a powerful tool for evidence-based clinical decision support.

## VIII. Conclusion

In this study, we addressed the critical challenge of preparing a specialized VLM for downstream clinical tasks by presenting and validating a complete pipeline for fine-tuning the MedGemma model to generate high-fidelity, structured captions for medical images across dermatology, fundus, and chest x-ray domains. By leveraging a novel dataset created through knowledge distillation and employing parameter-efficient fine-tuning techniques, we demonstrated substantial improvements in the model's classification performance compared to its baseline. Crucially, a rigorous evaluation using the RAGAS framework also confirmed significant gains in caption faithfulness and correctness, indicating the fine-tuned model's enhanced ability to generate reliable, factually grounded descriptions. The resulting model serves as a validated, high-quality query generator, producing captions with improved diagnostic accuracy and factual fidelity, thereby laying the essential groundwork for enhancing a multimodal RAG system to provide grounded, evidence-based clinical decision support from Malaysian CPGs.